\documentclass[11pt]{article}
\PassOptionsToPackage{table,dvipsnames,svgnames,x11names}{xcolor}
\usepackage{xcolor}

\usepackage[]{acl}
\usepackage{times}
\usepackage{colortbl} 
\usepackage{latexsym}
\usepackage[T1]{fontenc}
\usepackage[utf8]{inputenc}
\usepackage{microtype}
\usepackage{inconsolata}
\usepackage[addedmarkup=uwave,defaultcolor=red]{changes}
\usepackage{url}
\usepackage{tcolorbox}
\usepackage{graphicx}
\usepackage{makecell}
\usepackage{framed}
\usepackage{amsmath}
\usepackage{enumitem}
\usepackage{pifont}
\usepackage{wrapfig}
\usepackage{booktabs}
\usepackage{amsfonts}       
\usepackage{nicefrac}       
\usepackage{microtype}      
\usepackage{makecell}
\usepackage{multirow}
\usepackage[capitalize]{cleveref}
\usepackage{scalerel}
\usepackage[misc]{ifsym}
\usepackage{float}
\usepackage{stfloats}
\usepackage{subcaption} 
\usepackage{ulem}
\usepackage{fix-cm}
\makeatletter
\DeclareRobustCommand\onedot{\futurelet\@let@token\@onedot}
\def\@onedot{\ifx\@let@token.\else.\null\fi\xspace}

\makeatother

\makeatletter
\renewcommand{\paragraph}{%
  \@startsection{paragraph}{4}%
  {\z@}{0ex \@plus 0ex \@minus 0ex}{-1em}%
  {\normalfont\normalsize\bfseries}%
}
\makeatother

\newcommand{\yang}[1]{\textcolor{OliveGreen}{[Yang] #1}}

\definecolor{shadecolor}{rgb}{0.92,0.92,0.92}
\definecolor{bblue}{HTML}{0000fd}
\definecolor{rred}{HTML}{fe0000}
\definecolor{ggreen}{HTML}{018201}
\definecolor{redd}{HTML}{680020}
\definecolor{reddd}{HTML}{b71a36}
\definecolor{redddd}{HTML}{ebb9a7}

\crefname{figure}{Fig.}{Figs.}
\crefname{table}{Tab.}{Tabs.}
\crefname{section}{Sec.}{Secs.}

\title{\pipeline: Improving Meta Introspection of Small LLMs \\ by Learning Self-Reflection }

\author{
  Jiaqi Li\textsuperscript{1},
  Xinyi Dong\textsuperscript{2},
  Yang Liu\textsuperscript{1}, 
  Zhizhuo Yang\textsuperscript{2},
  Quansen Wang\textsuperscript{1,2},
  \\
  Xiaobo Wang\textsuperscript{1},
  Songchun Zhu\textsuperscript{1,2},
  Zixia Jia\textsuperscript{1,\corref{cor1},\dag},
  Zilong Zheng\textsuperscript{1,\corref{cor2},\dag}
  \\
  \textsuperscript{1}State Key Laboratory of General Artificial Intelligence, BIGAI \\
  \textsuperscript{2}Peking University
  \\
  \correspondence{
    \corref{cor1}Correspondence to Zixia Jia \texttt{<jiazixia@bigai.ai>} \\
    \corref{cor2}Correspondence to Zilong Zheng \texttt{<zlzheng@bigai.ai>}
  }
  \\
  \textsuperscript{\dag}These authors contributed equally to this work
}

\author{
  Jiaqi Li\textsuperscript{1},
  Xinyi Dong\textsuperscript{2},
  Yang Liu\textsuperscript{1}, 
  Zhizhuo Yang\textsuperscript{2},
  Quansen Wang\textsuperscript{1,2},
  \\
  \textbf{
  Xiaobo Wang\textsuperscript{1},
  Songchun Zhu\textsuperscript{1,2},
  Zixia Jia\textsuperscript{1, \Letter},
  Zilong Zheng\textsuperscript{1, \Letter}
  }
\\
 \textsuperscript{1}State Key Laboratory of General Artificial Intellligence, BIGAI \\
 \textsuperscript{2}Peking University
\\
 \small{
   \texttt{\{lijiaqi, jiazixia, zlzheng\}@bigai.ai}}
}

\usepackage{xspace}
\newcommand{\pipeline}{\textbf{ReflectEvo}\xspace}
\newcommand{\ds}{\textbf{ReflectEvo-460k}\xspace}

\begin{document}
\maketitle

\begin{abstract}

We present a novel pipeline \pipeline to demonstrate that small language models (SLMs) can enhance meta introspection through reflection learning. This process iteratively generates self-reflection for self-training, fostering a continuous and self-evolving process. Leveraging this pipeline, we construct \ds, a large scale, comprehensive self-generated reflection dataset with broadened instructions and diverse multi-domain tasks. Building upon this dataset, we demonstrate the effectiveness of \textbf{reflection learning} to improve SLMs' reasoning abilities using SFT and DPO with remarkable performance, substantially boosting Llama-3 from 52.4\% to 71.2\% and Mistral from 44.4\% to 71.1\%. It validates that ReflectEvo can rival or even surpass the reasoning capability of the three prominent open-sourced models on BIG-bench without distillation from superior models or fine-grained human annotation. We further conduct a deeper analysis on the high quality of self-generated reflections and their impact on error localization and correction. Our work highlights the potential of continuously enhancing the reasoning performance of SLMs through iterative reflection learning in the long run.

\end{abstract}

\section{Introduction}

Self-reflection involves meditating on, examining, and evaluating one’s behaviors, thoughts, motivations, and desires~\citep{atkins1993reflection, von1992reflections, denton2011reflection}. Typically, it inspects the reasoning process leading to the current solution, identifies errors in each step, generates critiques on the causes of the failure, and offers advice for refining the solution to improve the problem-solving performance of Large Language Models (LLMs)~\citep{welleck2022generating, ferraz2024llm,li2024loogle, wu2024cream}. Unlike the paradigm of learning directly from the reasoning process and final answer, we refer to it as the process of human-like \textit{meta introspection}, which explicitly generates self-reflection, providing textual differentiation and gradients as clear critiques and guidance on what to learn and how to improve based on the current state.

Recent research has demonstrated that LLMs can self-improve through their intrinsic capability of self-reflection~\citep{huang2022large, renze2024self,guo2025deepseek,wang2024exovip}. However, conventional approaches rely closely on LLMs with large model sizes or supervision distilled from a superior model. In this study, we challenge whether \textit{the self-reflection capability of SLMs can be learned effectively from reflection data.} However, it usually requires high-cost on fine-grained human annotation to acquire high-quality data for fine-tuning and is impractical to scale. Therefore, we are also curious whether \textit{it is possible to effectively utilize both high- and low-quality self-generated data from weaker models for reflection learning.} With this in mind, we aim to investigate the effectiveness of reflection learning via self-training~\citep{luong2024reft, qu2024recursive, pang2023language, tang2024mars} and further validate that the improvement of self-reflection can further strengthen LLM's inherent reasoning capabilities across various methods and tasks with more interpretability and generalization. We believe that this paradigm can act as a plug-and-play enhancement for various reasoning methods, which emulates human learning through a slower and deeper thought process that iteratively and ultimately derives self-evolution~\citep{li2023reflection, he2024self, li2024ram, tang2024mars, tang2023icsr}. 

\begin{figure*}[t!]
    \centering\includegraphics[width=1\linewidth]{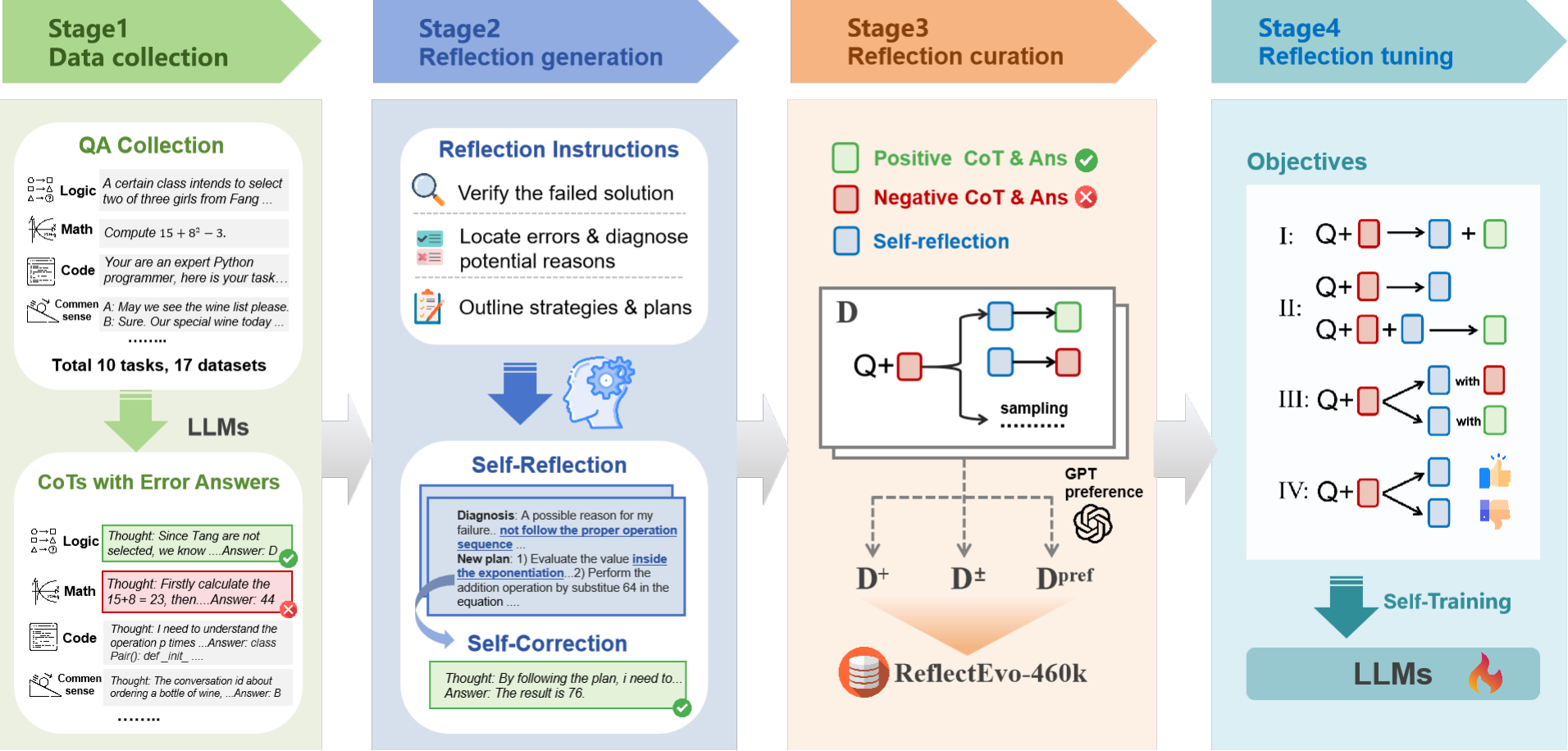}
    \caption{Overview pipeline of \pipeline.}
    \label{fig:pipeline}
    \vspace{-8pt}
\end{figure*}

Therefore, in this paper, we propose a novel pipeline \pipeline (\cref{ReflectEvo-460k}), to automatically generate self-reflection data and leverage self-training to enhance LLM's reflection capability. To the best of our knowledge, we are the first to demonstrate the potential of \textbf{meta introspection} of LLMs that are asked to explicitly generate reflection as an intermediate step-by-step process supervision rather than directly mapping an initial solution to a revised solution. 

Building on this pipeline, we curate a large-scale, diverse, and unsupervised reflection learning dataset \ds containing 460k reflection samples derived from 17 source datasets spanning 10 tasks and domains. We explore the diversity of reflection instructions and bootstrap multiple comparative reflections conditioned on the same question and initial solution. 
Based on the data, we develop \textbf{reflection learning} (\cref{reflection_tuning}) to further improve the self-reflection and self-correction capabilities of LLMs. 

The evaluation results validate the effectiveness of reflection learning in boosting the reasoning of weak models. It shows significant improvements on Llama-3, exceeding the original base model by 10\% on average tasks and outperforming its strongest counterpart with model size $\times 8$. We conduct a deeper analysis of the self-generated reflection data including various error types identified from the reflection and observe their gains on corrected answers.

In summary, our main contributions are: 
\begin{itemize}[leftmargin=*, topsep=0pt, noitemsep]
    
    \item  \textbf{Novel Pipeline for Self-Reflection Generation}: We propose ReflectEvo for automatic self-reflection generation and curation, which is the first to explore meta introspection of SLMs.
    \item  \textbf{Large-Scale and Diverse Self-generated Reflection Dataset}: We curate a comprehensive reflection training set ReflectEvo-460K from multiple data sources and tasks including various reflection instructions and comparative samples.
    \item  \textbf{Learning Reflection Via Self-training}: We develop four settings of reflection learning methods to effectively improve self-reflection and self-correction based on SFT and DPO, which significantly boost the reasoning abilities of SLMs as well as surpassing their stronger counterparts.
\end{itemize}

\section{The \pipeline Generation Scheme}\label{ReflectEvo-460k}
In this section, we introduce the end-to-end pipeline \pipeline for collecting self-generated reflections as the training data for \cref{reflection_tuning}, leveraging the inherent ability of SLMs (see \cref{{fig:pipeline}}).

\subsection{Problem Definition and Prelinminary}

Given a question $q$ and its ground truth answer $a^*$, the answer of the LLM after reasoning is denoted as $a$ followed by its corresponding verbal feedback $f$ from the environment, where $f$ represents the evaluation function that assesses whether an answer is correct or incorrect by comparing it to the reference answer $a^*$. The self-reflection $r$ of an LLM explicitly locates and analyzes errors in $a$ and makes further plans to mitigate the errors. Based on $r$ and the context provided in the previous stage, the LLM is then asked to revise its original answer $a$ to obtain $\hat{a}$ and solve $q$ as correctly as possible.



\subsection{Reflection Generation}\label{sec:generation}

\noindent\textbf{Step 1: Collection of instruction pool}\quad{}
To enhance the effectiveness and quality of the generated reflections $r$, we design instructions that target three key stages of reflection and correction, as defined below: 
\textbf{i. Verify the failed solution.} It analyzes the initial solution by tracing and examining the reasoning process with or without step-by-step verification. 
\textbf{ii. Locate errors and diagnose potential reasons.} It points out errors in specific reasoning steps and identifies the causes~\citep{zeng2024mr}. We delicately design prompts to mitigate the most common error types~\citep{li2024fb}, including mathematical (calculation \& algorithm) errors, logic and reasoning errors (flawed rationale \& internal inconsistency), instruction violation (context misinterpretation, incomplete or irrelevant response \& format discrepancy), factual errors. They are explicitly specified in the instructions for error elimination and accurate fault localization.
\textbf{iii. Outline strategies and plans for error correction and mitigation.} It provides strategies and guidance to address the error by proposing a high or low-level plan to mitigate similar issues in the future.

\begin{table*}[t]
\centering
\setlength{\tabcolsep}{4pt}
\resizebox{1\linewidth}{!}
{
\begin{tabular}{lcccccccccccc}
\toprule 
\textbf{Feature} & \parbox{2cm}{\centering \textbf{Logical} \\ \textbf{Reasoning}} & \parbox{2cm}{\centering \textbf{Mathematics}} & \parbox{2cm}{\centering \textbf{Coding}}  & \parbox{2cm}{\centering \textbf{Contextual} \\ \textbf{QA}} & \parbox{2.5cm}{\centering \textbf{Context-Free} \\ \textbf{QA}} & \parbox{2.5cm}{\centering \textbf{Reading} \\ \textbf{Comprehension}} & \parbox{2.5cm}{\centering \textbf{Commonsense} \\ \textbf{Reasoning}} & \parbox{2cm}{\centering \textbf{Social} \\ \textbf{Reasoning}} & \parbox{2cm}{\centering \textbf{Causal} \\ \textbf{Reasoning}} & \parbox{2cm}{\centering \textbf{Physics} \\ \textbf{Reasoning}} & \parbox{2cm}{\centering \textbf{Total}} \\ \midrule 
\# Reflection training samples & 253,405 & 92,967 & 9,125 & 19,399 & 3,624 & 32,135 & 22,044 & 19,175 & 8,757 & 1,168 & 461,799\\
\# Q\&A-Reflection samples & 164,746 & 106,434  & 7,520 & 17,448 & 2,940 & 20,760 & 10,012 & 11,404 & 3,212 & 468 & 344,944 \\
\% Correct after reflection & 16.60 & 10.77 & 15.31 & 4.26 & 9.66 & 12.39 & 33.88 & 19.99 & 39.98 & 38.46 & 13.93 \\
\# Avg. question length & 140 & 77 & 99 &  335  & 200    & 148  & 219 & 52 & 118 & 39 & 130 \\ 
\# Avg. answer length (turn 1) & 131 & 267 &  187 & 91 & 118  & 82 & 158 & 110 & 118 & 112 & 189 \\ 
\# Avg. answer length (turn 2) & 163 & 299 & 202 & 119 & 135 & 116  & 159 & 133 & 130 & 124 & 237 \\ 
\# Avg. reflection length & 238 & 222 & 254 & 267 & 251 & 261 & 268 & 252 & 259 & 256 & 250 \\ 
\bottomrule 
\end{tabular}}
\caption{Statistics of \textbf{\ds}. The average length in the table is computed by tokens.} 
\label{tab:statistics}
\vspace{-8pt}
\end{table*}

\noindent\textbf{Step 2: Data generation}\quad{} Based on the instructions outlined in Step 1, we introduce two components for reflection generation: a \textbf{Generator $G$} (reasoning model) that generates the initial answer with its reasoning process and a \textbf{Reflector $R$} (reflection model) that improves the incorrect answer through self-reflection and self-correction.   

\noindent\underline{\textit{Generator $G$}} Given a $q$, $G$ is built upon a base LLM instructed to generate interleaved thoughts and an initial answer $G(a|q)$. It is implemented as described in ReAct~\citep{yao2022react}, as the first step for self-reflection. We obtain the external environment feedback $f$ by evaluating the correctness of $a$ as a verifier. $f$ is a binary signal ``correct/incorrect'' with limited information, which is usually the case in real scenarios, eliminating the need for enriched feedback from humans or more powerful models. If correct, $a$ is directly used as the final answer. If incorrect, $R$ is used to revise the solution iteratively. In this paper, we perform self-reflection once to maximize the efficiency of self-generated data; however, this approach can be extended to multiple iterations in future studies.

\noindent\underline{\textit{Reflector $R$}}  We use exactly the same base LLM as $G$ for $R$. The generation process for $R$ is decomposed into \textbf{two phases: self-reflection and self-correction.} Self-reflection generates $R(r|q,a,f)$ to identify errors in the reasoning process and conduct a deeper analysis of the causes. Self-correction refines $a$ as $R(\hat{a}|q,a,f,r)$. To enrich the self-training data, we sample $k$ solution $\{r_j, \hat{a}_j\}_{j=1}^k$ for each $\{q, a, f\}$ conditioned on one specific prompt using reject sampling~\citep{liu2023statistical} to enrich the self-training data. We vary the prompts selected from the instruction pool to generate diverse self-reflection samples.

\subsection{Reflection Curation} ~{}
\\
After the above-mentioned process, we obtain a reflection training set with $M (N*k*m)$ samples:
{\small
    \begin{equation}
        \mathcal{D} = \{q_i, a_i, f_i, (r_{i,j}, \hat{a}_{i,j})_{j=1}^{k*m}\}_{i=1}^{N},
    \end{equation}
}
where $N$ is the number of QA pairs in $\mathcal{D}$, $m$ is the number of reflection instructions from pool, and $k$ is the value of reject sampling. We aim to further curate the data for reflection learning as follows.

First, we filter $r$ to include those followed by the correct $\hat{a}$, indicating that these reflections are of high quality for error correction, denoted as $\mathcal{D}^+$:
{\small
    \begin{equation}
        \mathcal{D}^+ = \{ (q_i, a_i, f_i, r_i,\hat{a}_i) \mid (\hat{a}_i = a^*) \}_{i=1}^{\left|\mathcal{D}\right|}
    \end{equation}
}

Subsequently, we leverage GPT-4o~\citep{hurst2024gpt} as a stronger teacher model to further select preferred reflection data from $\mathcal{D}^+$ to create pairwise data, denoted as $\mathcal{D}^\text{pref}$:
{\small
    \begin{equation}
        \mathcal{D}^\text{pref} =\{ (q_i, a_i, f_i, [y^{\text{cho}}_i,y^{\text{rej}}_i])\mid \exists \,y^{\text{cho}}_i,y^{\text{rej}}_i\}_{i=1}^{\left|\mathcal{D}^+\right|},
    \end{equation}
}
where $y=(r,\hat{a})$ is the reflection and corresponding corrected answer. $y^{cho}$ and $y^{rej}$ are solutions randomly selected for each $\{q, a, f\}$ whose $r$ is chosen and rejected, respectively, by GPT-4o.

To fully utilize low-quality reflection data followed by $\hat{a}_i$ that is still judged to be incorrect, we enrich the self-training data by incorporating both positive and negative samples as pairwise data for each $\{q, a, f\}$, denoted as \textbf{$\mathcal{D}^\pm$}. 
{\small
    \begin{equation}
        \mathcal{D}^\pm =\{ (q_i, a_i, f_i, [y^+_i,y^-_i])\mid \exists \, y^+_i,y^-_i\}_{i=1}^{\left|\mathcal{D}\right|},
    \end{equation}
}
where $y^+$ and $y^-$ are solutions  whose $\hat{a}$ is evaluated as correct or incorrect by $a^*$.


\begin{figure*}[ht]
    \centering
    \begin{subfigure}[t]{0.35\textwidth}
        \centering
        \includegraphics[width=0.95\linewidth]{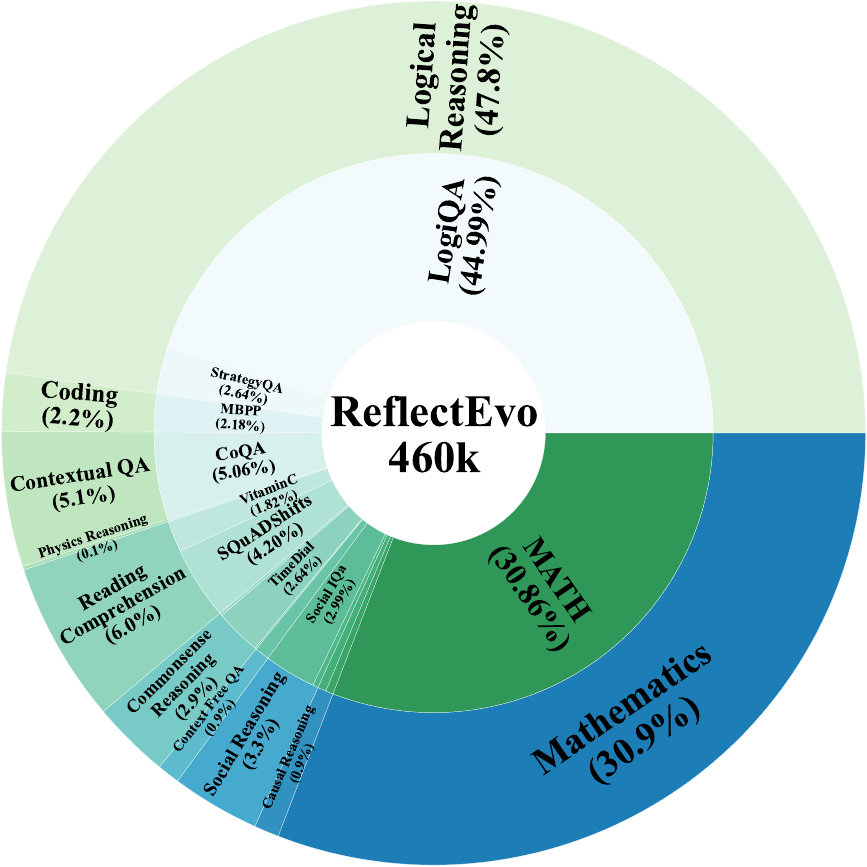}
        \caption{}
        \label{fig:distribution}
    \end{subfigure}
    \hfill
    \begin{subfigure}[t]{0.63\textwidth}
        \centering
        \includegraphics[width=\linewidth]{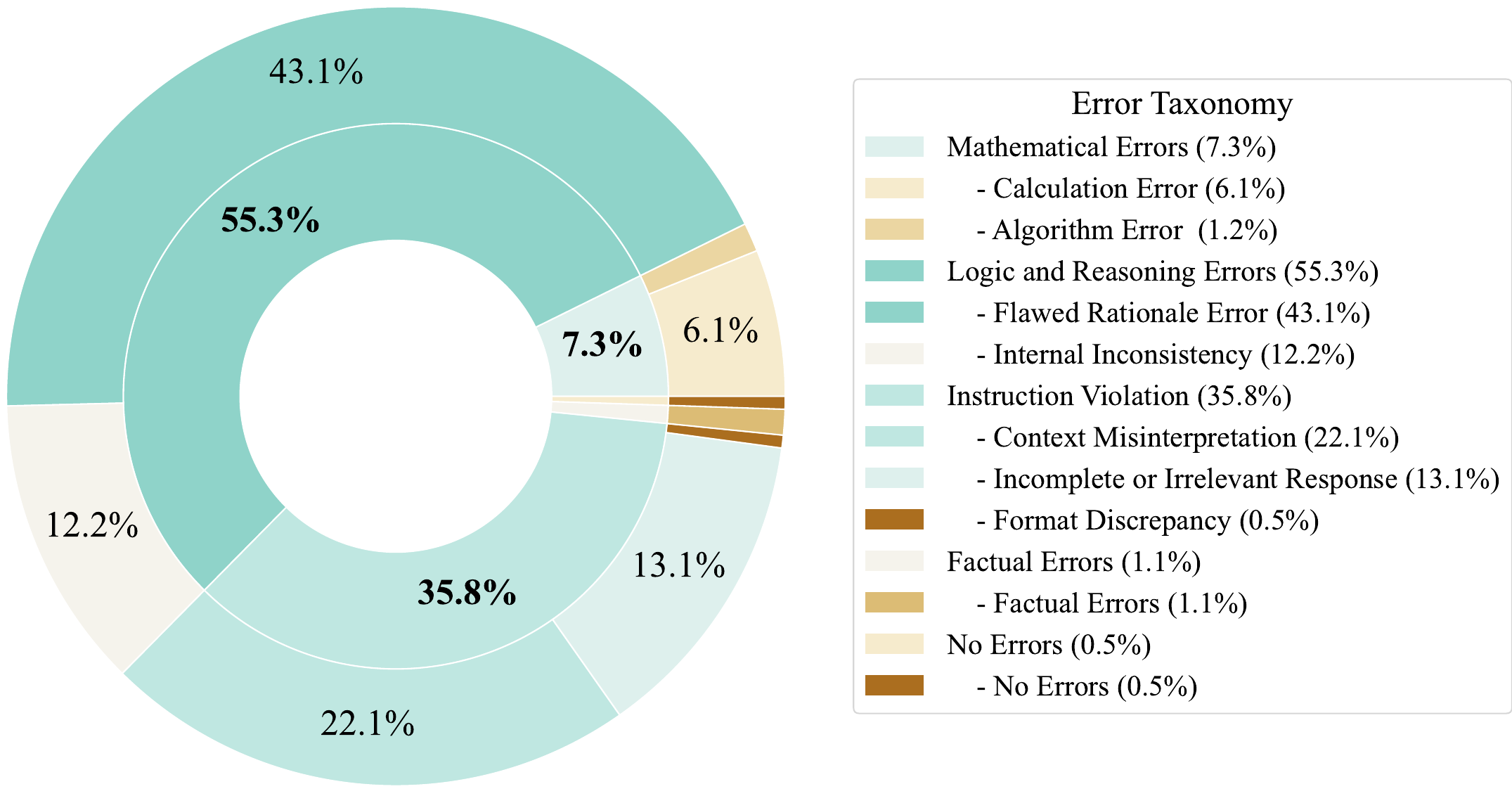}
        \caption{}
        \label{fig:error_type_distribution}
    \end{subfigure}
    \caption{
        (a) Task-dataset hierarchy distribution of ReflectEvo-460k.
        (b) Error type distribution of corrected thoughts identified by reflection in the test sets. 
    }
    \label{fig:combined_figures}
\end{figure*}

Following the above-mentioned steps in \cref{ReflectEvo-460k}, We create a reflection training dataset \ds by curating examples of 17 carefully selected source subsets from LogiQA~\citep{liu2020logiqa}, MATH~\citep{hendrycksmath2021}, MBPP~\citep{austin2021program}, and BIG-bench~\citep{srivastava2023beyond}, spanning diverse domains and categories. The Statistics of the dataset are shown in \cref{tab:statistics} and \cref{{fig:distribution}}. We use three commonly used SLMs including Llama-3-8B~\citep{dubey2024llama}, Mistral-7B~\citep{jiang2023mistral7b} and Gemma-2-9B~\citep{team2024gemma} for the entire process of data generation, training, and test. Implementation details and instructions are provided in \cref{appendix:prompts,appendix:reflectiongen_details}.

\section{Reflection Learning on Self-Generated Data}\label{reflection_tuning}
In this section, we further investigate the effectiveness of reflection learning on the reflector $R$ by adopting self-training on \ds using supervised fine-tuning (SFT)~\citep{ouyang2022training} and Direct Preference Optimization (DPO)~\citep{rafailov2024direct}.

\subsection{Reflection Learning}

We use \textbf{SFT} on $\mathcal{D}^+$ in two different settings below. This strengthens the model to better leverage reflections as intermediate thoughts leading to positive $\hat{a}$ for refinement. 

\textbf{Setting 1:} We train the capacity of self-reflection and self-correction in one stage:
{\small
    \begin{equation}\label{eq:objective1}
        \mathcal{L}_1 = -\mathbb{E}_{(q, a, f, r, \hat{a})\sim \mathcal{D^+}} \log R((r,\hat{a}) \mid q, a, f)
    \end{equation}
}

\textbf{Setting 2:} We train the capacity of self-reflection and self-correction respectively in two stages:

{\small
    \begin{equation}\label{eq:objective2_1}
        \mathcal{L}_{2.1} = - \mathbb{E}_{(q, a, f, r)\sim \mathcal{D^+}} \log R (r\mid q, a, f) 
    \end{equation}
}
{\small
    \begin{equation}\label{eq:objective2_2}
        \mathcal{L}_{2.2} = - \mathbb{E}_{(q, a, f, r, \hat{a})\sim \mathcal{D^+}} \log R (\hat{a}\mid q, a, f, r)
    \end{equation}
}

Inspired by error-driven learning from humans,  we also leverage negative samples $\hat{a}^-$ that comprise a large portion of $\mathcal{D}^\pm$ and offer valuable insights for model enhancement. In addition, we assume GPT-4o with better self-reflection, which is required for reflection preference annotation as $\mathcal{D}^\text{pref}$, guiding SLMs to continuously refine reflections. We use preference learning through \textbf{DPO} on the aforementioned pairwise data to better judge and distinguish high-quality reflections from suboptimal ones in the following settings.

\textbf{Setting 3:} We train self-reflection only on $\mathcal{D}^\pm$:
{\small
    \begin{equation}\label{eq:objective3}
        \mathcal{L}_3 = - \mathbb{E}_{(x, r^+, r^-)\sim \mathcal{D^\pm}} \log \sigma [r_\theta(x, r^+) - r_\theta(x, r^-)]
    \end{equation}
}
{\small
    \begin{equation}\label{eq:objective3_1}
        r_\theta(x, r) = \beta \log \frac{\pi_\theta(r\mid x)}{\pi_\text{ref}(r\mid x)},  x = (q, a, f)
    \end{equation}
}




\textbf{Setting 4:} We train self-reflection only on $\mathcal{D}^\text{pref}$:
{\small
    \begin{equation}\label{eq:objective4}
        \mathcal{L}_4 = - \mathbb{E}_{(x, r^\text{cho}, r^\text{rej})\sim \mathcal{D^\text{pref}}}\log \sigma [r_\theta(x, r^\text{cho}) - r_\theta(x, r^\text{rej})]
    \end{equation}
}
where $R(\cdot)$ is the policy model $\pi_\theta$ and $G(\cdot)$ is the reference model $\pi_\text{ref}$. $\sigma$ is the logistic function and $\beta$ is a hyperparameter that controls the proximity to the reference policy $G(\cdot)$ in both settings 3 and 4. The objective is to steer $R(\cdot)$ towards increasing the likelihood of $r^+$ with the correct solutions $\hat{a}$ or chosen $r^{\text{cho}}$ and decreasing the likelihood of $r^-$ with incorrect solutions $\hat{a}$ or rejected $r^{\text{rej}}$ for given $(q,a,f)$. More details can be found in \cref{appendix:training_details}. 

\subsection{Inference}
During inference, the process follows the same steps as those of reflection data generation in \cref{sec:generation}. We use the model after reflection learning as a reflector at the inference time for self-reflection and correction. It can be implemented as a multi-turn rollout that terminates either when the current $a$ is judged to be correct or when it reaches the predefined maximum number of turns (two turns in our setting using twice QA with one intermediate reflection). 


\section{Experiments}

\subsection{Performance Learning on \pipeline}
We measure the performance of self-reflection by adopting the following metrics: 1) \textbf{Acc@t1}: the model’s accuracy in the first turn; 2) \textbf{Acc@t2}: the model’s accuracy in the second turn; 3) \textbf{$\Delta$(t1,t2)}: accuracy improvement between the first and second turns measuring the efficacy of self-reflection. 

We compare three main methods in our experiments, including prompt-based QA with or without reflection without training, SFT training with direct answers, and self-training based reflection learning introduced in \cref{reflection_tuning} from Setting 1 to Setting 4, noted as one-stage w/ $\mathcal{D}^+$, two-stage w/ $\mathcal{D}^+$, w/ $\mathcal{D}^\pm$ and w/ $\mathcal{D}^\text{pref}$ respectively. 

\paragraph{Overall Performance on Different Tasks}
\cref{tab:llama_main} illustrates the overall performance on ReflectEvo. We discard the self-generated reflection data by Mistral on MATH due to its extremely low quality. We observe that LLMs gain more from prompt-based reflection, whereas SLMs show either minor improvements or degradation. This is primarily because without specialized training, SLMs inherently generate low-quality reflections and fail to leverage feedback effectively for self-correction. For comparison, experiments on our self-training methods show significant improvements in both models and various tasks. Specifically, it achieves over 20\% in $\Delta$(t1,t2) for Llama-3 on MBPP and BIG-bench as well as Mistral on LogiQA and BIG-bench. Notably, all three evaluated models outperform their stronger model using ReflectEvo on BIG-bench. This indicates that different models and tasks benefit greatly from the four self-training methods, even surpassing the SFT on answers without step-by-step reasoning process, which paves the way for broader applications and scenarios for various SLMs. 

\begin{figure}[t!]
    \centering
    \includegraphics[width=\linewidth]{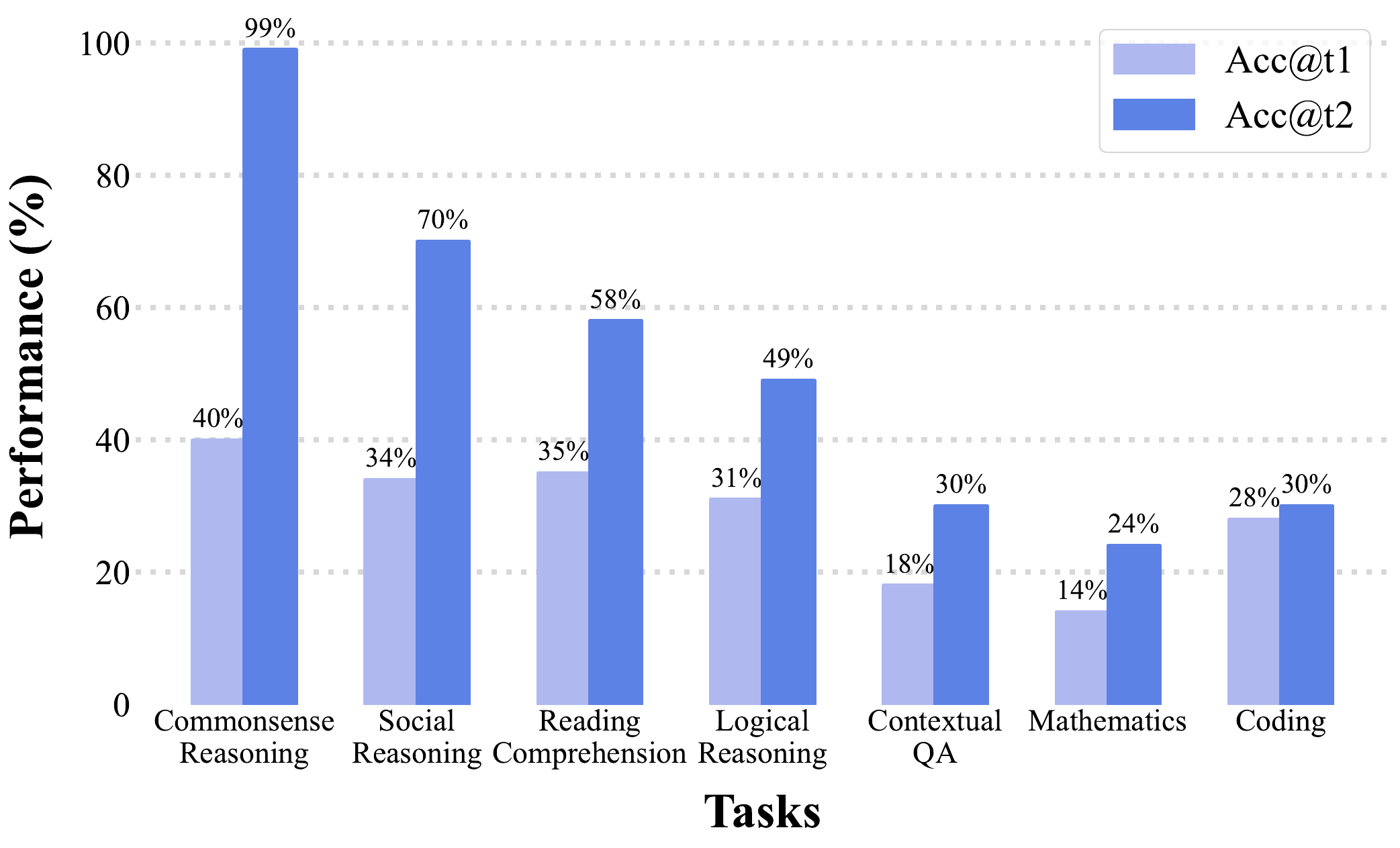}
    \caption{Performance training with ReflectEvo across different tasks on Llama-3-8B.}
    \label{fig:task_paradigms}
\end{figure}

\cref{fig:task_paradigms} provides an in-depth analysis on the reflection learning across tasks. Our method significantly contributes to various tasks, including reasoning, math, QA, and comprehension, with an average of 22\% in $\Delta$(t1,t2). For coding, it only improves to a certain degree probably due to the lack of fine-grained step-by-step critiques on the erroneous solutions for reflection training on models that are not specialized in coding. 

\begin{table*}[t]
  \centering
  \resizebox{\textwidth}{!}{
  \begin{tabular}{lcccccccccccc}
  \toprule
  & \multicolumn{3}{c}{\textbf{LogiQA}} & \multicolumn{3}{c}{\textbf{MATH
   }}  & \multicolumn{3}{c}{\textbf{MBPP}}  & \multicolumn{3}{c}{\textbf{BIG-bench}}\\
  \cmidrule(lr){2-4} \cmidrule(lr){5-7}  \cmidrule(lr){8-10} \cmidrule(lr){11-13}
   & \textbf{Acc@t1} & \textbf{Acc@t2} & \textbf{$\Delta$(t1,t2)} & \textbf{Acc@t1} & \textbf{Acc@t2} & \textbf{$\Delta$(t1,t2)}  & \textbf{Acc@t1} & \textbf{Acc@t2} & \textbf{$\Delta$(t1,t2)} & \textbf{Acc@t1} & \textbf{Acc@t2} & \textbf{$\Delta$(t1,t2)} \\
  \midrule
  \multicolumn{13}{c}{\cellcolor{lightgray!50}\textbf{Meta-Llama-3-8B-Instruct}}  \\
  \midrule
   \textbf{Prompt based} &  &  &  &   &   &   &  &   &  \\
 \hspace{0cm}$\hookrightarrow$ w/o reflection & \multirow{2}{*}{30.2\%} & 38.8\% & +8.6\% & \multirow{2}{*}{14.4\%} & 15.0\% &
+0.6\%  & \multirow{2}{*}{28.4\%} & 44.0\% & +15.6\% & \multirow{2}{*}{38.2\%} & 52.4\% & +14.2\% \\
 \hspace{0cm}$\hookrightarrow$ w/ reflection &  & 36.2\% & +6.0\% &  & 16.0\% & +1.6\%
  &  & 45.8\% & +17.4\% &  & 51.0\% & +12.8\%\\
  \midrule
  \textbf{SFT based} & &  &  &   &   &   &  &   &  \\
  \hspace{0cm}$\hookrightarrow$ w/ SFT qa pairs & 46.6\% & -  & - & 10\%  & - & - 
  & 31.2\% & - & - & 61.6\% & - & -\\
  \midrule
  \textbf{Self-training based (Ours)} &  &  &  &   &   &   &  &   &  \\
  \hspace{0cm}$\hookrightarrow$ one-stage w/ $\mathcal{D}^+$  & \multirow{4}{*}{30.2\%}  & \cellcolor{blue!10}\textbf{43.8\%} &  \cellcolor{blue!10}\textbf{+13.6\%} & \multirow{4}{*}{14.4\%} &  \cellcolor{blue!10}\textbf{23.6\%} & \cellcolor{blue!10}\textbf{+9.2\%}
  & \multirow{4}{*}{28.4\%} & 29.6\% & +1.2\% & \multirow{4}{*}{38.2\%} &  \cellcolor{blue!10}\textbf{71.2\%} & \cellcolor{blue!10}\textbf{+33.0\%}  \\
 \hspace{0cm}$\hookrightarrow$ two-stage w/ $\mathcal{D}^+$ &  & \cellcolor{blue!10}\textbf{49.4\%} & \cellcolor{blue!10}\textbf{+19.2\%} &  & 14.5\% & +0.1\%
  &  & 42.4\% & +14.0\% & & 45.4\% & +7.2\% \\
  \hspace{0cm}$\hookrightarrow$ w/ $\mathcal{D}^\pm$  & &\cellcolor{blue!10}\textbf{41.8\%}  &\cellcolor{blue!10}\textbf{+11.6\%}  &  &15.2\% &+0.8\% &  &\cellcolor{blue!10}\textbf{48.8\%}\  &\cellcolor{blue!10}\textbf{+20.4\%}   & &\cellcolor{blue!10}\textbf{63.0\%} &\cellcolor{blue!10}\textbf{+24.8\%}  \\
  \hspace{0cm}$\hookrightarrow$ w/ $\mathcal{D}^\text{pref}$  &  &\cellcolor{blue!10}\textbf{39.2\%}  &\cellcolor{blue!10}\textbf{+9.0\%}  &  &14.8\% &+0.4\% &  &\cellcolor{blue!10}\textbf{47.4\%}  &\cellcolor{blue!10}\textbf{+19.0\%} & &\cellcolor{blue!10}\textbf{59.6\%} &\cellcolor{blue!10}\textbf{+21.4\%} \\
 \midrule
 \multicolumn{13}{c}{\cellcolor{lightgray!50}\textbf{Meta-Llama-3-70B-Instruct}}  \\
 \midrule
  \hspace{0cm}$\hookrightarrow$ w/o reflection & \multirow{2}{*}{42.4\%} &64.4\%  &+22.0\%  & \multirow{2}{*}{40.8\%} &49.6\% &+8.8\%   & \multirow{2}{*}{66.2\%}  &71.0\%  &+4.8\% & \multirow{2}{*}{48.0\%} &67.0\% &+19.0\% \\
  \hspace{0cm}$\hookrightarrow$ w/ reflection & &56.8\%  &+14.4\%  & &48.6\% &+7.8\%   &  &73.0\%  &+6.8\% & &64.4\% &+16.4\% \\
  \midrule
  \midrule
  \multicolumn{13}{c}{\cellcolor{lightgray!50}\textbf{Mistral-7B-Instruct-v0.2}}  \\
  \midrule
   \textbf{Prompt based} &  &  &  &   &   &   &  &   &  \\
  \hspace{0cm}$\hookrightarrow$ w/o reflection & \multirow{2}{*}{28.8\%} & 31.2\% & +2.4\% & \multirow{2}{*}{9.2\%} & 10.6\% & +1.4\%  & \multirow{2}{*}{20.4\%} & 23.0\% & +2.6\% & \multirow{2}{*}{36.6\%} & 43.8\% & +7.2\% \\
  \hspace{0cm}$\hookrightarrow$ w/ reflection &  & 34.2\% & +5.4\% &   & 10.2\% & +1.0\% &   & 23.6\% & +3.2\% &  & 44.4\% & +7.8\%\\
  \midrule
  \textbf{SFT based} &  &  &  &   &   &   &  &   &  \\
  \hspace{0cm}$\hookrightarrow$ w/ SFT qa pairs & 28.8\% & - & - & 7.6\%  & -&-
  & 17.0\% & - & - & 37.8\% & - & - \\
  \midrule
  \textbf{Self-training based (Ours)} &  &  &  &   &   &   &  &   &  \\
  \hspace{0cm}$\hookrightarrow$ one-stage w $\mathcal{D}^+$  & \multirow{4}{*}{28.8\%}  &\cellcolor{blue!10}\textbf{38.4\%}  &\cellcolor{blue!10}\textbf{+9.6\%}  & - & - & -
  &\multirow{4}{*}{20.4\%}  &\cellcolor{blue!10}\textbf{24.0\%} &\cellcolor{blue!10}\textbf{+3.6\%} &\multirow{4}{*}{36.6\%} &\cellcolor{blue!10}\textbf{51.6\%} &\cellcolor{blue!10}\textbf{+15.0\%} \\
  \hspace{0cm}$\hookrightarrow$ two-stage w/ $\mathcal{D}^+$  & & \cellcolor{blue!10}\textbf{48.8\%} & \cellcolor{blue!10}\textbf{+20.0\%} & - & - & - &  & 20.8\% & +0.4\%   &   & \cellcolor{blue!10}\textbf{71.1\%}  &  \cellcolor{blue!10}\textbf{+34.5\%} \\
   \hspace{0cm}$\hookrightarrow$ w/ $\mathcal{D}^\pm$  &  & \cellcolor{blue!10}\textbf{39.2\%} & \cellcolor{blue!10}\textbf{+10.4\%}   & - & - & -  &  &23.2\%  &+2.8\%  &  &\cellcolor{blue!10}\textbf{50.2\%}  &\cellcolor{blue!10}\textbf{+13.6\%}\\
   \hspace{0cm}$\hookrightarrow$ w/ $\mathcal{D}^\text{pref}$  &  &\cellcolor{blue!10}\textbf{38.0\%}  &\cellcolor{blue!10}\textbf{+9.2\%}   & - & - & -  & &22.6\%  &+2.2\%  & &\cellcolor{blue!10}\textbf{48.4\%}  &\cellcolor{blue!10}\textbf{+11.8\%}\\
 \midrule
 \multicolumn{13}{c}{\cellcolor{lightgray!50}\textbf{Mistral-22B-Small-Instruct}}  \\
 \midrule
 \hspace{0cm}$\hookrightarrow$ w/o reflection &\multirow{2}{*}{46.4\%} &62.8\%  &+16.4\% &\multirow{2}{*}{47.4\%} &56.2\% &+8.8\%   &\multirow{2}{*}{63.0\%}  &68.0\%  &+5.0\% &\multirow{2}{*}{54.4\%}  &67.2\%   &+12.8\% \\
  \hspace{0cm}$\hookrightarrow$ w/ reflection & &62.0\%  &+15.6\% & &52.8\% &+5.4\%   &  &68.0\%  &+5.0\% & &68.0\%   &+13.6\% \\
  \bottomrule
  \end{tabular}
  }
  \caption{Performance on Llama-3 and Mistral using ReflectEvo.}
   \label{tab:llama_main}
  \end{table*}



\begin{figure}[t!]
    \centering
    \includegraphics[width=\linewidth]{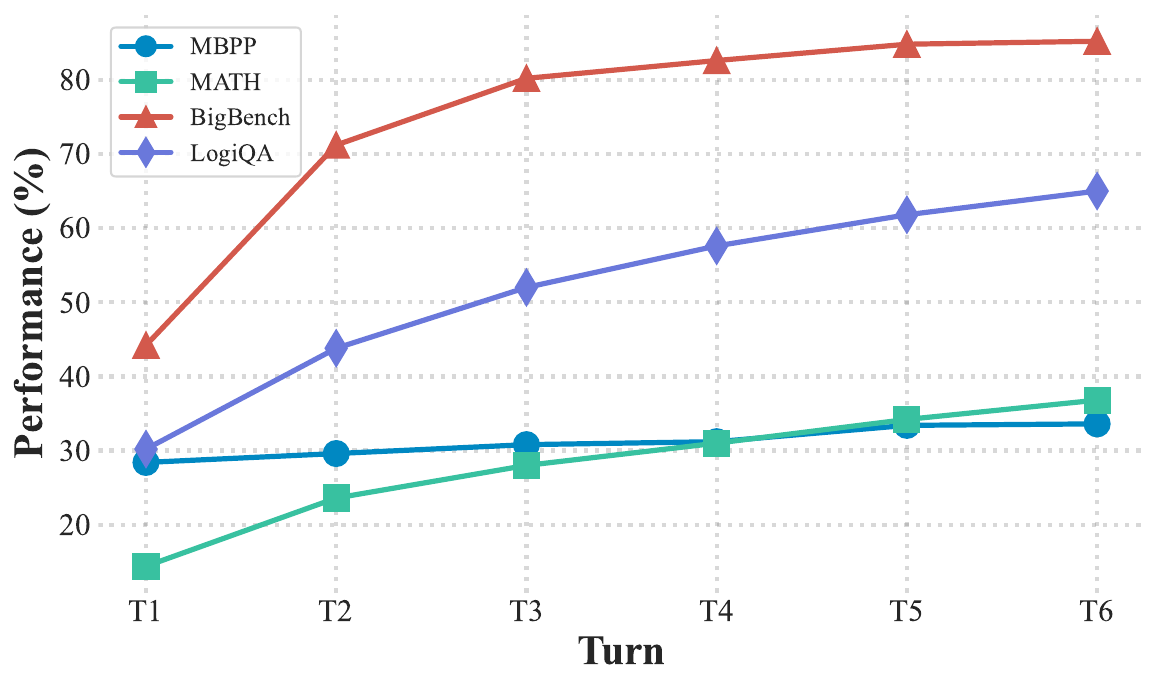}
    \caption{Performance in multi-turn self-reflection on Llama-3-8B after tuning.}
    \label{fig:multi_turn}
\end{figure}

\paragraph{Effect of Reflection from Teacher Model}
To investigate the influence of reflection sources, we compare different reflections generated by the SLM itself and a more advanced model like GPT-4o which acts as a teacher model with greater knowledge and reasoning capabilities in \cref{tab:teacher_reflection}. Reflections from both models strengthen the $\Delta$(t1,t2) of QA performance after tuning under different settings proposed in \cref{reflection_tuning}, while the self-generated data require less cost and resources in practice. To our expectation, reflections from the teacher model yields more obvious improvements underscoring the benefits of high-quality reflection data generation and selection for further improvement.

\begin{table}[t!]
\centering
\resizebox{\linewidth}{!}{
\begin{tabular}{ll|cc}
\toprule
\textbf{Dataset} & \textbf{Method} & \textbf{Acc@t2}  & \textbf{$\Delta$(t1,t2)}  \\
\midrule
\multirow{5}{*}{\textbf{SR}} & prompt-based &36.2\% &+6.0\%  \\
& one stage w/ $\mathcal{D}^+$ & 43.8\% & +13.6\% \\
& two stage w/ $\mathcal{D}^+$  & \textbf{49.4\%} & \textbf{+19.2\%}  \\
& w/ $\mathcal{D}^\pm$  & 41.8\% & +11.6\% \\
& w/ $\mathcal{D}^\text{pref}$ & 39.2\% & +9.0\%\\
\midrule
 \multirow{5}{*}{\textbf{TR}}  & prompt-based &46.2\% &+16.0\%   \\
 & one stage w/ $\mathcal{D}^+$  & \textbf{52.0\%} & \textbf{+21.8\%}  \\
 & two stage w/ $\mathcal{D}^+$  & 41.2\% & +11.0\%  \\
 & w/ $\mathcal{D}^\pm$ & 48.8\% & +18.6\% \\
 & w/ $\mathcal{D}^\text{pref}$ & 48.0\% & +17.8\% \\ 
\bottomrule
\end{tabular}}
\caption{Performance on LogiQA using different sources of reflections on Llama-3 (Acc@t1=30.2\% from \cref{tab:llama_main}). \textbf{SR} and \textbf{TR} indicate self-reflection and teacher-reflection respectively.} 
\label{tab:teacher_reflection}
\end{table}

\paragraph{Scaling Multi-turn Self-reflection}
We further extend the application of self-reflection to multi-turn QA in \cref{fig:multi_turn}. To our expectation, the results demonstrate a consistent improvement with increasing turns of reflection on different tasks. BIG-bench exhibits the most significant improvement, surpassing 80\% accuracy after six turns and LogiQA also shows a notable upward trend, highlighting the effectiveness of iterative refinement. MBPP and MATH display relatively modest improvements with gradual increase, which suggesting that the impact of self-reflection learning is broadly beneficial but varies between tasks. It is encouraged to investigate the underlying factors that contribute to these differences to further enhance performance in various tasks.

\paragraph{Generalization across Different Tasks and Models} 
We conduct deeper studies on the generalization of the self-reflection capability after tuning across different tasks (\cref{tab:generalization_task}) and models (\cref{tab:generalization_model}). Our findings reveal that the benefits of reflection learning generalize across tasks, particularly for LogiQA and BIG-bench with 10\% increase, which commonly require strong reasoning abilities from LLMs. Due to the divergence of MATH and MBPP, there is merely improvement when trained on reflections generated from the other three datasets. We observe that all the test models in \cref{tab:generalization_model} benefit from the reflector after tuning for error correction in Acc@t2, especially for initial solutions from different generators. For Mistral and Gemma, even with a minor decrease compared with the corresponding results in \cref{{tab:llama_main}} and \cref{tab:gemma_main}, the result on these two models highlights the potential of our pipeline across different models and demonstrates the effectiveness of reflectors when applied to various generators.

In \cref{tab:generalization_rfl_data}, we further explore whether the self-reflection data of one LLM can be beneficial for the other. Compared with \cref{tab:llama_main} in our paper, we find that the reflection data generated by LlaMA-3.1-8B is helpful for Mistral-7B on reflection learning with comparable or even better performance. It indicates that our dataset \ds could be reusable for the community for future studies.

\begin{table*}[t]
  \centering
  \resizebox{\textwidth}{!}{
  \begin{tabular}{lcccccccccccc}
  \toprule
   & \multicolumn{3}{c}{\textbf{LogiQA}} & \multicolumn{3}{c}{\textbf{MATH
   }}  & \multicolumn{3}{c}{\textbf{BIG-bench}} & \multicolumn{3}{c}{\textbf{MBPP}} \\
  \cmidrule(lr){2-4} \cmidrule(lr){5-7}  \cmidrule(lr){8-10} \cmidrule(lr){11-13}
   & \textbf{Acc@t1} & \textbf{Acc@t2} & \textbf{$\Delta$(t1,t2)} & \textbf{Acc@t1} & \textbf{Acc@t2} & \textbf{$\Delta$(t1,t2)}  & \textbf{Acc@t1} & \textbf{Acc@t2} & \textbf{$\Delta$(t1,t2)} & \textbf{Acc@t1} & \textbf{Acc@t2} & \textbf{$\Delta$(t1,t2)} \\
  \midrule
  \textbf{Prompt based} &  &  &  &   &   &   &  &   &    &  &   & \\
  \hspace{0cm}$\hookrightarrow$ w/ reflection & 30.2\% & 36.2\% & +6.0\% & 14.4\% & 16.0\% & +1.6\% & 38.2\% & 51.0\% & +12.8\% & 28.4\% & 45.8\% & +17.4\% \\
  \midrule
  \textbf{Self-training based (Ours)} &  &  &  &   &   &   &  &   &    &  &   & \\
  \hspace{0cm}$\hookrightarrow$ w/ $\mathcal{D}^+$ on LogiQA  & \multirow{4}{*}{30.2\%} & - & - & \multirow{4}{*}{14.4\%}   &14.4\% &+0.0\%  &  \multirow{4}{*}{38.2\%} &\cellcolor{blue!10}\textbf{60.0\%}  &\cellcolor{blue!10}\textbf{+21.8\%} &\multirow{4}{*}{28.4\%}  &30.6\%   &+2.2\% \\
  \hspace{0cm}$\hookrightarrow$ w/ $\mathcal{D}^+$ on MATH  &  &36.6\%  &+6.4\%  &  & - & - &  
  &\cellcolor{blue!10}\textbf{54.8\%}  &\cellcolor{blue!10}\textbf{+16.6\%}   &  &28.8\%   &+0.4\%\\
  \hspace{0cm}$\hookrightarrow$ w/ $\mathcal{D}^+$ on BIG-bench   &  &\cellcolor{blue!10}\textbf{52.0\%}  &\cellcolor{blue!10}\textbf{+21.8\%}  &   &14.4\% &+0.0\%  &  & - & -   &  &36.2\% &+7.8\%\\
  \hspace{0cm}$\hookrightarrow$ w/ $\mathcal{D}^+$ on MBPP  &  &30.4\%  &+0.2\%  &   &14.6\% &+0.2\%  &  &40.2\%  &+2.0\%     &   & -  & -\\
  
  \bottomrule
  \end{tabular}
  }
  \caption{Generalization across tasks for Llama-3 training one-stage with different task-specific subsets in $\mathcal{D}^+$.}
  \label{tab:generalization_task}
  \vspace{-8pt}
  \end{table*}

\begin{table}[t]
\centering
\resizebox{\linewidth}{!}
{
\footnotesize
\begin{tabular}{l|ccc}
\toprule
\textbf{Different generators} & \textbf{Acc@t1} & \textbf{Acc@t2} & \textbf{$\Delta$(t1,t2)} \\
\midrule
Mistral-7B  &28.8\%  &45.8\% &+17.0\% \\
Gemma-2-9B  &47.6\%  &57.2\%  &+9.6\% \\
Llama-3.1-8B   &37.4\%  &51.0\%  &+13.6\% \\
\bottomrule
\end{tabular}}
\caption{Generalization across generators using the same reflector Llama-3 training one-stage with subset of LogiQA in $\mathcal{D}^+$. } \label{tab:generalization_model}
\vspace{-8pt}
\end{table}

\begin{table}[t!]
\centering
\resizebox{\linewidth}{!}{
\begin{tabular}{l|ccc}
\toprule
\textbf{Self-training method} & \textbf{Acc@t1} & \textbf{Acc@t2} & \textbf{$\Delta$(t1,t2)} \\
\midrule
one-stage w/ $\mathcal{D}^+$   & 28.8\% & \textbf{40.2\%} & \textbf{+11.4\%} \\ 
two-stage w/ $\mathcal{D}^+$    & 28.8\% & 38.0\% & +9.2\%  \\ 
w/ $\mathcal{D}^\pm$   & 28.8\% &39.4\%  & +10.6\% \\ 
w/ $\mathcal{D}^\text{pref}$    & 28.8\% & 38.6\% & +9.8\% \\ 
\bottomrule
\end{tabular}}
\caption{Generalization of generated reflection data on LogiQA of LlaMA-3.1-8B training on Mistral-7B}
\label{tab:generalization_rfl_data}
\end{table}

\paragraph{Effect of Different Verifiers on Self-Reflection} 

In this paper, self-reflection is performed only when the model's answer is verified as incorrect using the ground truth. Another potential approach is to train the model itself as a verifier or use an external reward function to score the model’s answer based on a predefined threshold. We compare the effects of using oracle ground truth and self-judgments generated by the baseline model as verifiers in \cref{tab:verifier}. For both verifiers, reflection learning improves Acc@t1 by an average of 13+\% and enhances Acc@t2 by up to 7\% compared with the untuned version. Although the baseline model, without specialized training, exhibits occasional misjudgments, its verification process results in minor performance degradation on the advantage of reflection learning. We leave this a direction for further exploration on the optimized verifiers in an end-to-end pipeline.


In \cref{tab:baselines}, we make further experiments by comparing with three well-acknowledged baselines for self-improvement through reflection or correction. To make the comparison fair, the external feedback used in all the experiments is only a binary signal ``correct/incorrect'' without further explanation or given ground truth. We follow the evaluation setting in the original paper. Comparing with STaR and Re-ReST, Ours benefit from improvement over turns. Our methods achieves much higher reasoning performance after self-reflection, which emulates human learning through a slower and deeper thought process that iteratively and ultimately derives self-evolution

\begin{table}[t]
\centering
\resizebox{\linewidth}{!}
{
\footnotesize
\begin{tabular}{l|cc}
\toprule
\textbf{Self-training Method} & \textbf{Oracle Groundtruth} & \textbf{Self-judgement} \\
\midrule
one-stage w/ $\mathcal{D}^+$   & 43.8\% & 32.8\%   \\
two-stage w/ $\mathcal{D}^+$   & 49.4\% & 50.2\%   \\
w/ $\mathcal{D}^\pm$  & 41.8\% &  40.6\%  \\
w/ $\mathcal{D}^\text{pref}$   & 39.2\% &  37.8\%  \\
\bottomrule
\end{tabular}}
\caption{Acc@t2 using different verifiers during inference on LogiQA for Llama-3. (Acc@t1=30.2\% and Acc@t2= 36.2\% without tuning from \cref{tab:llama_main})}
\label{tab:verifier}
\vspace{-8pt}
\end{table}

\begin{table}[t!]
\centering
\setlength{\tabcolsep}{3pt} 
\resizebox{0.9\linewidth}{!}{
\begin{tabular}{l|ccc}
\toprule
\textbf{Methods} & \textbf{Acc@t1} & \textbf{Acc@t2} & \textbf{$\Delta$(t1,t2)} \\
\midrule
STaR~\citep{zelikman2022star}   &  40.0\% & - & - \\ 
Re-ReST~\citep{dou2024re}  & 38.8\% & - & - \\ 
RISE~\citep{qu2024recursive}  &31.4\%  & 34.4\% & +3.0\% \\ 
one-stage w/ $\mathcal{D}^+$   & 30.2\% & 43.8\% & +13.6\% \\ 
two-stage w/ $\mathcal{D}^+$    & 30.2\% & \textbf{49.4\%} & \textbf{+19.2\%} \\ 
w/ $\mathcal{D}^\pm$   & 30.2\% & 41.8\% &  +11.6\%\\ 
w/ $\mathcal{D}^\text{pref}$    & 30.2\% & 39.2\%  & +9.0\% \\ 
\bottomrule
\end{tabular}}
\caption{Performance on different baselines using LlaMA-3.1-8B on LogiQA.}
\vspace{-10pt}
\label{tab:baselines}
\end{table}

\begin{figure*}[t!]
    \centering
    \includegraphics[width=1\linewidth]{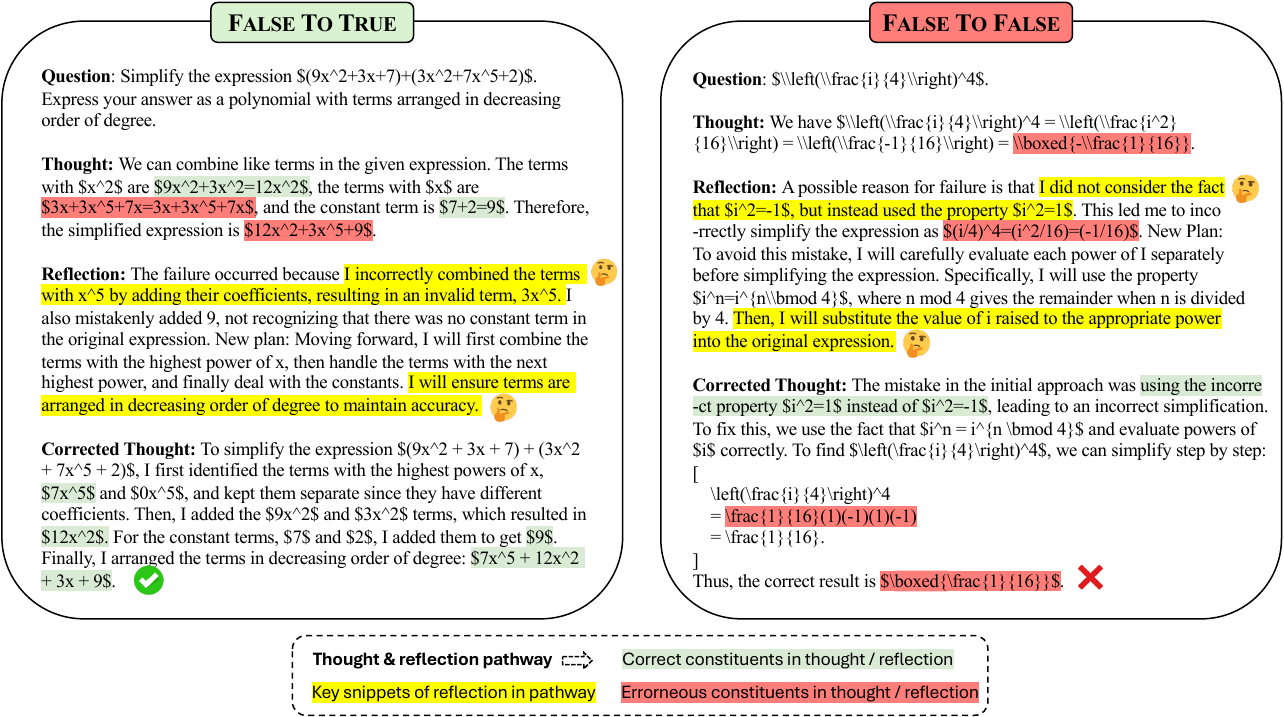}
    \caption{Qualitative examples from the MATH. ``False to True'' and a ``False to False'' stand for successful and failed correction in the second turn respectively. The key snippets highlighted in green, red and yellow indicate correct, erroneous thought and reflection respectively. }
    \label{fig:example}
    \vspace{-8pt}
\end{figure*}

\begin{figure}[t!]
    \centering
    \includegraphics[width=0.8\linewidth]{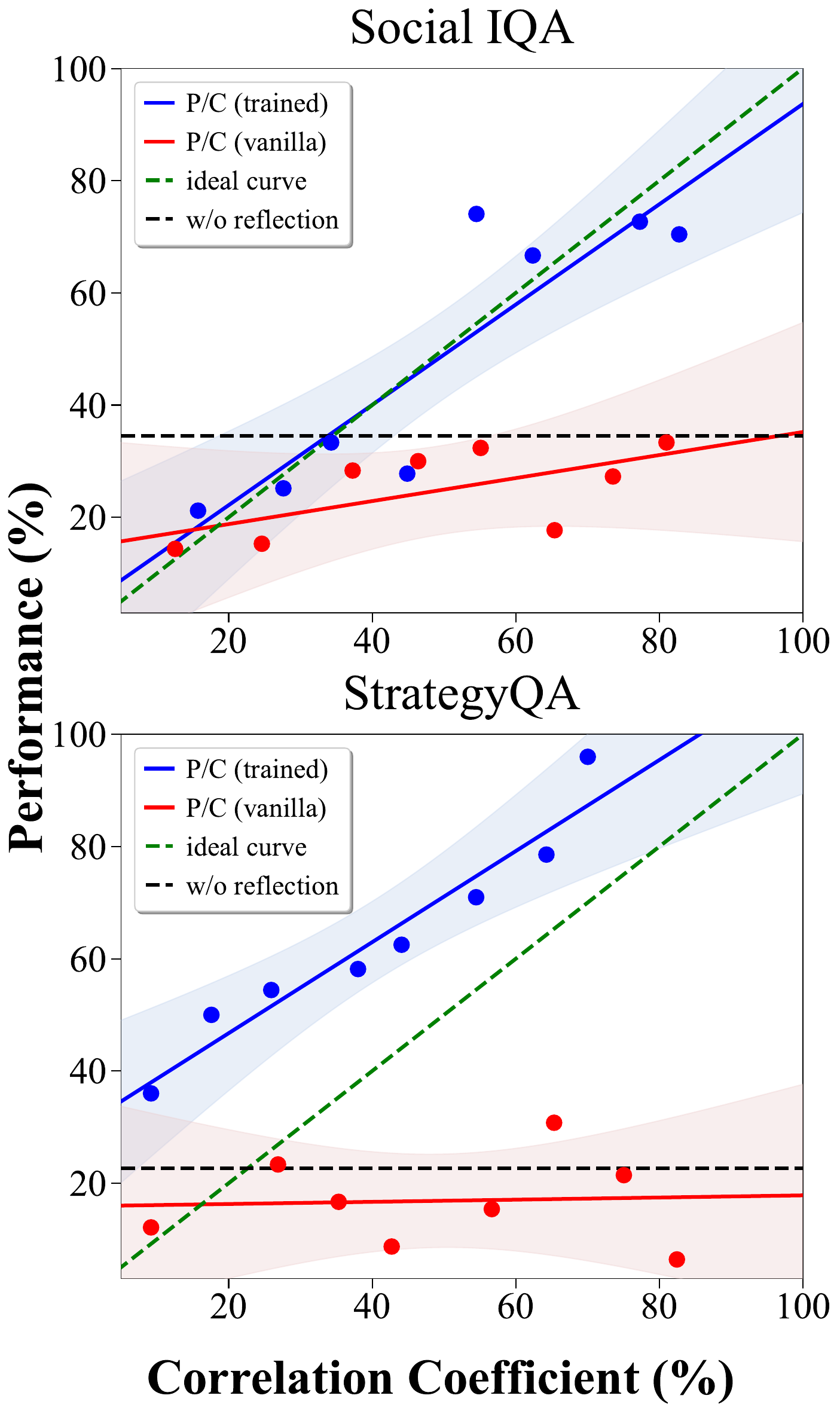}
    \caption{Task performance (Acc@t2) versus the correlation between reflection and the second-turn thought.}
    \label{fig:performance_gains_correlation}
    \vspace{-10pt}
\end{figure}

\subsection{In-depth Analysis on Reflection}
\label{sec:analysis}


\paragraph{Error Types Identified by Reflection}
To dissect the intrinsic properties of our reflection data, we analyze the error types in the initial thoughts specified by the reflection across all test sets. Potential error labels are generated heuristically by auto-tagging with GPT-4o and then calibrated by human annotators, achieving Cohen's kappa of 51.18\% with moderate agreement~\citep{landis1977measurement} indicating high annotation quality.

\cref{fig:error_type_distribution} shows five coarse-grained and nine fine-grained error types identified through human calibration. The most common errors are \textit{Logic and Reasoning Errors} (88.4\%) and \textit{Instruction Violation} (47.9\%), indicating that math and logic issues were the primary causes. We also provide detailed error distributions for the different subsets. MATH has a higher percentage of \textit{Calculation Errors} (20.8\%) than the other subsets, whereas COQA has more \textit{Context misinterpretation} (43.1\%). 
This shows that our method provides tailored reflections for specific domains rather than superficial or general advice.

\paragraph{Correlation Analysis in Reflection} 
We calculate the correlation between reflection and second-turn corrected thoughts, and we assess the association between the correlation and Acc@t2 after self-correction. Empirically, we hypothesize that they have a linear relationship, and we select the Pearson correlation coefficient by computing the semantic similarity for each pair of data (see the details in \cref{fig:stacked_perf_gains_correlation}). As we have seen, reflection learning can improve the ability of models to correct errors; we argue that if reflection is indeed specific to the error in thought, then task performance should intuitively be enhanced as the correlation between reflection and corrected thought increases. 

Measuring with the Pearson coefficient, \cref{fig:performance_gains_correlation} and \cref{fig:stacked_perf_gains_correlation} show that StrategyQA, Social IQa, VitaminC, and SQuAD all have a clear linear relationship between the performance and the correlation of reflection -- second-turn thought, while MATH and MBPP exhibit irrelevant tendency or show a slightly negative correlation implying their desire data of fine-grained reflection. Comparing the blue and red correlation curves, we find that more similarity between the reflection and corrected thought, more effective correction (\textit{i.e.}, higher performance) that outperforms the vanilla model. 

\paragraph{Case Studies} We perform case studies to see how reflection interacts with the thought process by making critiques and refinements in Fig. \ref{fig:example}. We random sampled 100 cases from the MATH test set and display two of them. In the case ``False to True'', reflection precisely recaps the key causes of error and explicitly bridges the logical pathway between the initial thought and the corrected one, which finally results in the correct answer. In contrast, we find that even tiny erroneous constituent in the reflection may lead to a false reasoning thought and final answer. It validates that high-quality reflection is helpful for incentivizing the model to generate thought with correct answer while flawed reflection still lead to repeated errors after self-correction, which aligns with similar findings on the impact of the reasoning steps in \citet{shinn2024reflexion} and \citet{zelikman2024quietstar}.


\section{Related Work}
\paragraph{Self-training and Self-Improvement}
Self-training allows a model to learn from its own outputs, reducing its reliance on human-annotated data or superior models~\citep{zelikman2022star, yuan2024self, chen2024self}. Previous research has primarily concentrated on enhancing models' reasoning abilities through SFT~\citep{yuan2023scaling} with positive samples or preference learning using both positive and negative samples to potentially leverage valuable information in incorrect solutions and recent advances also extend self-training to agentic scenarios~\citep{wang-etal-2024-self-training, wallace2024diffusion, gulcehre2023reinforced, song2024trial, motwani2024malt,li2024ram}. We further advocate reducing reliance on resource-heavy rationale annotations via self-training for SLMs.

\paragraph{Learning for Self-reflection}
Recent research highlights the significant benefits of integrating self-reflection into LLMs to enhance their reasoning and problem-solving capabilities, by iteratively refining their responses~\citep{kumar2409training, cheng2024vision, qu2024recursive, yao2023retroformer, zhou2024reflect, liang2024sheep, moskvoretskii2025self}. \citet{shinn2024reflexion} reinforces the language agent to verbally reflect on task feedback and induce better plans in subsequent trials. \citet{dou2024re} employs the low-quality outputs generated from the weak model iteratively by fine-tuning the reflection module for self-refinement. \citet{zhang2024small} further validates that SLMs have the ability of self-correction on reasoning tasks by accumulating high-quality critique-correction data. We pioneer the exploration of reflection learning on self-generated data.

\section{Conclusion}
We propose ReflectEvo to enhance SLMs through reflection learning by iteratively generating self-training data, which achieves substantial performance improvements, even surpassing much larger models highlighting its generalization across various models and tasks for future research.


\section*{Limitations}
Despite the promising results of \pipeline through reflection learning, there are several limitations to our work. Firstly, the quality of the self-generated reflection data is highly dependent on the initial reasoning ability of the SLMs. Models with inherently weak reasoning capabilities may struggle to produce high-quality reflections, which in turn limits the effectiveness of the self-training process. Secondly, while our pipeline demonstrates significant improvements in certain tasks, tasks such as coding and mathematics require more specialized knowledge and step-by-step critiques than reasoning and comprehension tasks. Additionally, future work could explore more sophisticated feedback mechanisms with optimized verifiers or reward functions during inference to enhance the reflection learning process. Addressing these limitations is crucial for further advancing the self-reflection capabilities of SLMs.

\section*{Ethics Statement} 
We adhere to ethical principles to ensure the responsible development and application of our proposed techniques. Our work focuses on enhancing the self-reflection abilities of SLMs without directly involving human subjects or sensitive information. We acknowledge the potential broader impacts of our research, recognize the environmental and computational costs associated with LLM training, and strive to optimize our methods for efficiency. 
 
\bibliography{custom}

\appendix

\onecolumn

\section{Further analyais and results}
\cref{tab:gemma_main} indicates that BIG-bench gains more from reflection tuning with Acc@t2 = 78.4\% with a substantial increase of +17.2\% compared with other tasks and the baseline methods. However, our method on Gemma-2 shows marginal improvement compared with Llama-3 and Mistral probably due to its inherent strong reasoning capability (comparable performance on different models in sizes of 9B \& 27B). It may either need selection of higher-quality reflection data or supervision from superior models and further optimization on the training methods for reflection enhancement. Due to the defect of SFT training without step-by-step reasoning process and the limited number of training data, the SFT performance of MATH and MBPP degrade due to the nature of fast thinking than thoses with reflection.

  \begin{table*}[t]
  \centering
  \resizebox{\textwidth}{!}{
  \begin{tabular}{lcccccccccccc}
  \toprule
   & \multicolumn{3}{c}{\textbf{LogiQA}} & \multicolumn{3}{c}{\textbf{MATH
   }}  & \multicolumn{3}{c}{\textbf{MBPP}}  & \multicolumn{3}{c}{\textbf{BIG-bench}}\\
  \cmidrule(lr){2-4} \cmidrule(lr){5-7}  \cmidrule(lr){8-10} \cmidrule(lr){11-13}
   & \textbf{Acc@t1} & \textbf{Acc@t2} & \textbf{$\Delta$(t1,t2)} & \textbf{Acc@t1} & \textbf{Acc@t2} & \textbf{$\Delta$(t1,t2)}  & \textbf{Acc@t1} & \textbf{Acc@t2} & \textbf{$\Delta$(t1,t2)} & \textbf{Acc@t1} & \textbf{Acc@t2} & \textbf{$\Delta$(t1,t2)} \\
  \midrule
  \multicolumn{13}{c}{\cellcolor{lightgray!50}\textbf{Gemma-2-9B-it}}  \\
  \midrule
   \textbf{Prompt based} &  &  &  &   &   &   &  &   &  \\
  \hspace{0cm}$\hookrightarrow$ w/o reflection  &\multirow{2}{*}{47.6\%} &  63.0\% &  +15.4\% & \multirow{2}{*}{34.6\%} &  40.4\% &  +5.8\%  & \multirow{2}{*}{54.4\%} & 59.2\% & +4.8\% & \multirow{2}{*}{61.2\%} & 75.6\% & +14.4\% \\
  \hspace{0cm}$\hookrightarrow$ w/ reflection &  & 60.0\% & +12.4\% &  & 40.0\% & +5.4\% &  &  59.6\% &  +5.2\% &  & 74.4\% & +13.2\%\\
  \midrule
  \textbf{SFT based} & &  &  &   &   &   &  &   &  \\
  \hspace{0cm}$\hookrightarrow$ w/ SFT qa pairs & 50.6\% & - & - & 18.6\% & - &- 
  & 37.0\% & - & - & 74.6\% & - & - \\
  \midrule
  \textbf{Self-training based (Ours)} &  &  &  &   &   &   &  &   &  \\
  \hspace{0cm}$\hookrightarrow$ one-stage w/ $D^+$  &\multirow{4}{*}{47.6\%}  &62.4\%  &+14.8\%  &\multirow{4}{*}{34.6\%}  &40.0\% &+5.4\%
  &\multirow{4}{*}{54.4\%}  &57.8\%  &+3.4\%  &\multirow{4}{*}{61.2\%} & \cellcolor{blue!10}\textbf{78.4\%} & \cellcolor{blue!10}\textbf{+17.2\%}\\
  \hspace{0cm}$\hookrightarrow$ two-stage w/ $D^+$  &  & 60.6\% & +13.0\% &  & 35.0\% & +0.4\%
  &  & 56.8\% & +2.4\% & & 67.0\% & +5.8\%  \\
  \hspace{0cm}$\hookrightarrow$ w/ $D^\pm$  &  &62.6\%  &+15.0\%  &  &40.0\% &+5.4\%  &  &58.6\%  &+4.2\%  & &74.8\% &+13.6\% \\
  \hspace{0cm}$\hookrightarrow$ w/ $D^\text{pref}$  &  & 63.0\%  & +15.4\%  & &40.0\% &+5.4\% &  &59.4\% &+5.0\%  &  &75.0\%  &+13.8\%  \\
 \midrule
 \multicolumn{13}{c}{\cellcolor{lightgray!50}\textbf{Gemma-2-27B-it}}  \\
 \midrule
  \hspace{0cm}$\hookrightarrow$ w/o reflection & \multirow{2}{*}{52.0\%} & 59.2\% & +7.2\% & \multirow{2}{*}{38.4\%} & 43.8\% & +5.4\%  &\multirow{2}{*}{65.4\%}  & 69.4\%  & +4.0\%  &\multirow{2}{*}{63.4\%} &72.0\% & +8.6\% \\
  \hspace{0cm}$\hookrightarrow$ w/ reflection &  &  65.4\% & +13.4\% &  & 45.0\% &  +6.6\% & &69.2\% &+3.8\% & &  75.2\%  & +11.8\%  \\
  \bottomrule
  \end{tabular}
  }
  \caption{Performance on Gemma-2 using ReflectEvo.}
   \label{tab:gemma_main}
  \end{table*}


\section{Implementation Details}  
\subsection{Reflection generation}\label{appendix:reflectiongen_details}
In this paper, we conduct the self-reflection once during the process of the two turns of reasoning and answering for both data generation and inference across most experiments. The generalization performance of multi-turn self-reflection can be found in \cref{fig:multi_turn}.

The number of reject sampling $k$ is 2. To validate the effectiveness of reflection tuning on various tasks, we incorporate 14 datasets selected from BIG-bench besides LogiQA, MATH and MBPP. Those datasets are delicately selected to focus more on the comprehension and reasoning abilities across diverse domains, comprising a comprehensive collection of dataset. The datasets includes: Commonsense Reasoning (RiddleSense, TimeDial, Known Unknowns), Social Reasoning (Social IQa, Implicit Interpersonal Relations), Reading Comprehension (VitaminC, SQuADShifts), Logical Reasoning (StrategyQA, Analytic Entailment), Contextual QA (CoQA Conversational Question Answering), Context Free QA (Truthful QA), Causal Reasoning (Causal Judgment, Cause and Effect), and Physics Reasoning (Physical Intuition). For datasets with more than 1000 samples, we randomly select 1000 QA pairs; for datasets with fewer than 1000 samples, we retain the entire original set. 

Each reflection instruction consists of the three stages introduced in step 1 in \cref{ReflectEvo-460k} and different variants of prompts used in each stage can be seen \cref{appendix: rfl_prompt}. The combination of them forms a diverse, comprehensive instruction pool with 32 (2*8*2) instructions used in step 2. For each dataset, we random select 5 or 6 of the instructions ($M$) to generate the reflections in \ds considering the data generation efficiency. 

For each task, we use corresponding subset for training without using the whole \ds. For example, we use the training set of LogiQA for reflection generation and learning for LlaMA-3.1-8B and evaluate the same model on the test set of LogiQA in the experiments.

\begin{table}[t]
  \centering
  \resizebox{0.8\textwidth}{!}{
  \begin{tabular}{lcccccccccccc}
  \toprule
   & \multicolumn{3}{c}{\textbf{LogiQA}} & \multicolumn{3}{c}{\textbf{MATH
   }}  & \multicolumn{3}{c}{\textbf{BIG-bench}} & \multicolumn{3}{c}{\textbf{MBPP}} \\
  \cmidrule(lr){2-4} \cmidrule(lr){5-7}  \cmidrule(lr){8-10} \cmidrule(lr){11-13}
   & \textbf{$D^+$} & \textbf{$D^\pm$} & \textbf{$D^\text{pref}$} & \textbf{$D^+$} & \textbf{$D^\pm$} & \textbf{$D^\text{pref}$}  & \textbf{$D^+$} & \textbf{$D^\pm$} & \textbf{$D^\text{pref}$} & \textbf{$D^+$} & \textbf{$D^\pm$} & \textbf{$D^\text{pref}$} \\
  \midrule
  \textbf{Training} &25371  &152475  &59870  &13796  &50946   &28225 &20410  &70672   &30909  &1151  &5365    &2609    \\
  \textbf{Testing} &  &500  &  &   &500   &   &  &500   &    &  &500   & \\
  \bottomrule
  \end{tabular}
  }
  \caption{Data statistics for training and testing samples in the experiments.}
  \label{tab:training_testing}
  \end{table}

\begin{table}[ht]
    \centering
    \resizebox{0.7\textwidth}{!}{\begin{tabular}{lcccc}
    \toprule
    Hyperparameter & one-stage w/ $D^+$  & two-stage w/ $D^+$  &  w/ $D^\pm$ &  w/ $D^\text{pref}$ \\
    \midrule
    learning rate & 1e-3 & 1e-5,1e-3 &5e-7, 7e-7 & 5e-7, 7e-7 \\
    weight decay & 0 & 0-0.01 & - & - \\
    max grad norm & 1.0 & 1.0 & - & - \\
    $\beta_{1}$ for SFT & 0.9 & 0.99 & - & - \\
    $\beta_{2}$ for SFT & 0.999 & 0.9 & - & - \\
    $\beta$ for DPO & - & - & 0.01 & 0.01 \\
    $\epsilon$ & 1e-8 & 1e-08 & -  & - \\
    max new tokens & 512 & 512 & 248 & 248 \\
    \bottomrule
    \end{tabular}}
    \caption{The hyperparameters for reflection tuning.}
    \label{tab:hyperparameter}
    \vspace{-8pt}
\end{table}

\subsection{Training}
\label{appendix:training_details}
All the experiments for reflection tuning can be conducted on two Nvidia A100 80GB GPU, 32GB memory, 128 Core AMD CPU. The resource costs are mainly dependent on the tuning methods (full-parameter fine-tuning, parameter-efficient fine-tuning (PEFT) and DPO), the sizes of the models, and the sizes of the datasets. The main hyperparameters used for different settings of reflection tuning are shown in \cref{tab:hyperparameter}. The learning rate varies based on different models and tasks.

For one stage training with w/ $D^+$, we use LoRA-based PEFT in this setting with 4-bit quantization via BitsAndBytes. We set LoRA rank $r=8$, scaling factor $\alpha=32$, and a dropout rate of $0.1$. The per-device batch size is set to 16 for LogiQA and 8 for others. For two stage training with w/ $D^+$, we use the full-parameter SFT in this setting with bfloat16 data precision. The per-device batch size is set to 16 with gradient accumulation of 4.
For DPO training with both \( D^\pm \) and \( D^\text{pref} \), the per-device training batch size is set to 2, and gradient accumulation is set to 32.

\begin{figure*}[ht]
    \centering
    \includegraphics[width=0.9\linewidth]{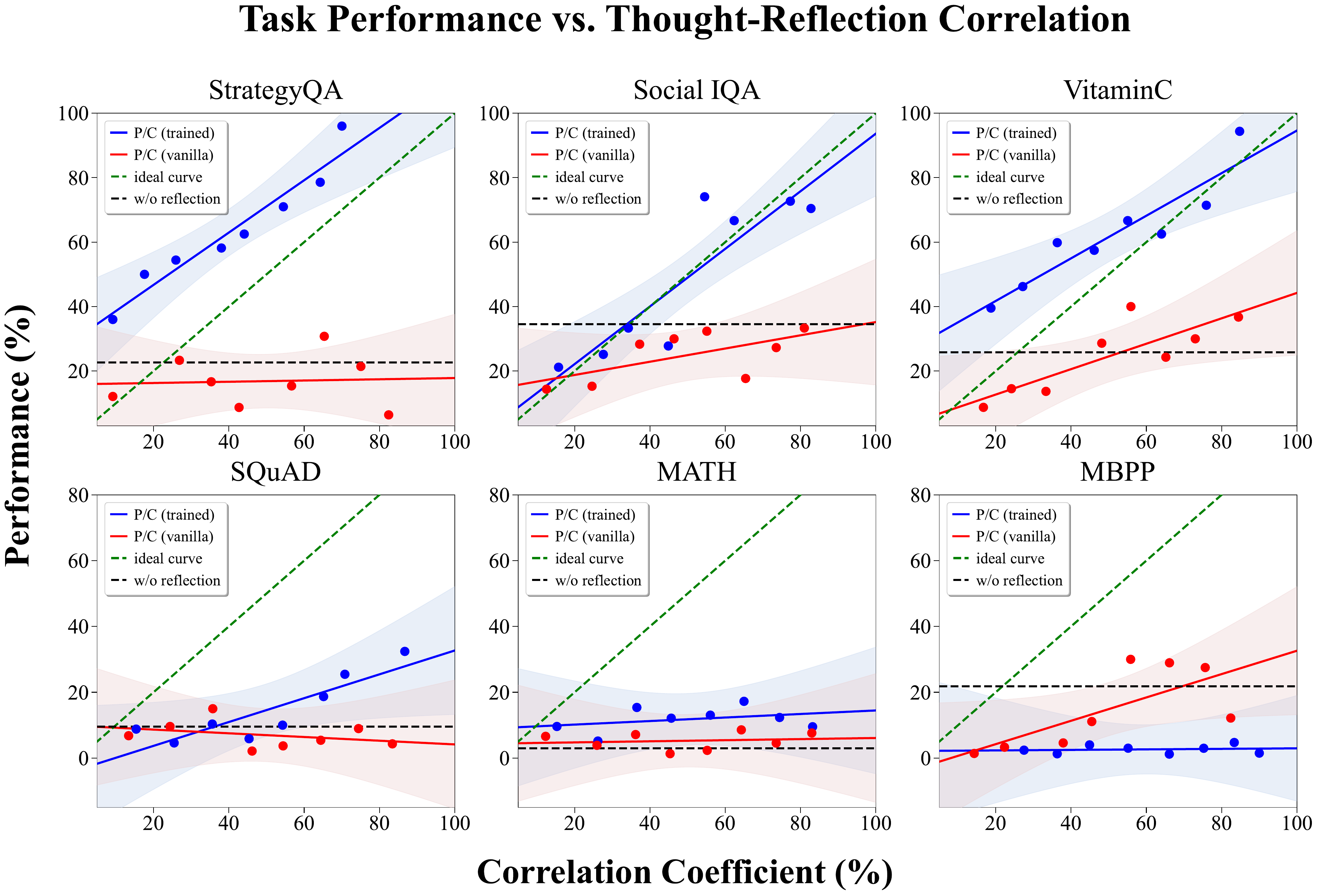}
    \caption{Task performance (Acc@t2) versus the correlation between reflection and second-turn thoughts for Llama-3-8B with self-training reflection (\textcolor{bblue}{blue \dotuline{dots} and \uline{curve}}) and prompt-based reflection (\textcolor{rred}{red \dotuline{dots} and \uline{curve}}). The ideal correlation (\textcolor{ggreen}{\dashuline{green dashed curve}}) denotes a standard linear tendency for comparison purposes, and the \dashuline{black dashed line} represents Llama-3-8B without reflection. Note that the y-values of the spots denote the average performance (axis-y), where an array of test data points is located in a specific interval of the correlation coefficient (axis-x), and the correlation coefficient of these spots is also averaged by the values in the same interval.}
    \label{fig:stacked_perf_gains_correlation}
\end{figure*}

\begin{figure*}[ht]
    \centering
    \includegraphics[width=0.7\linewidth]{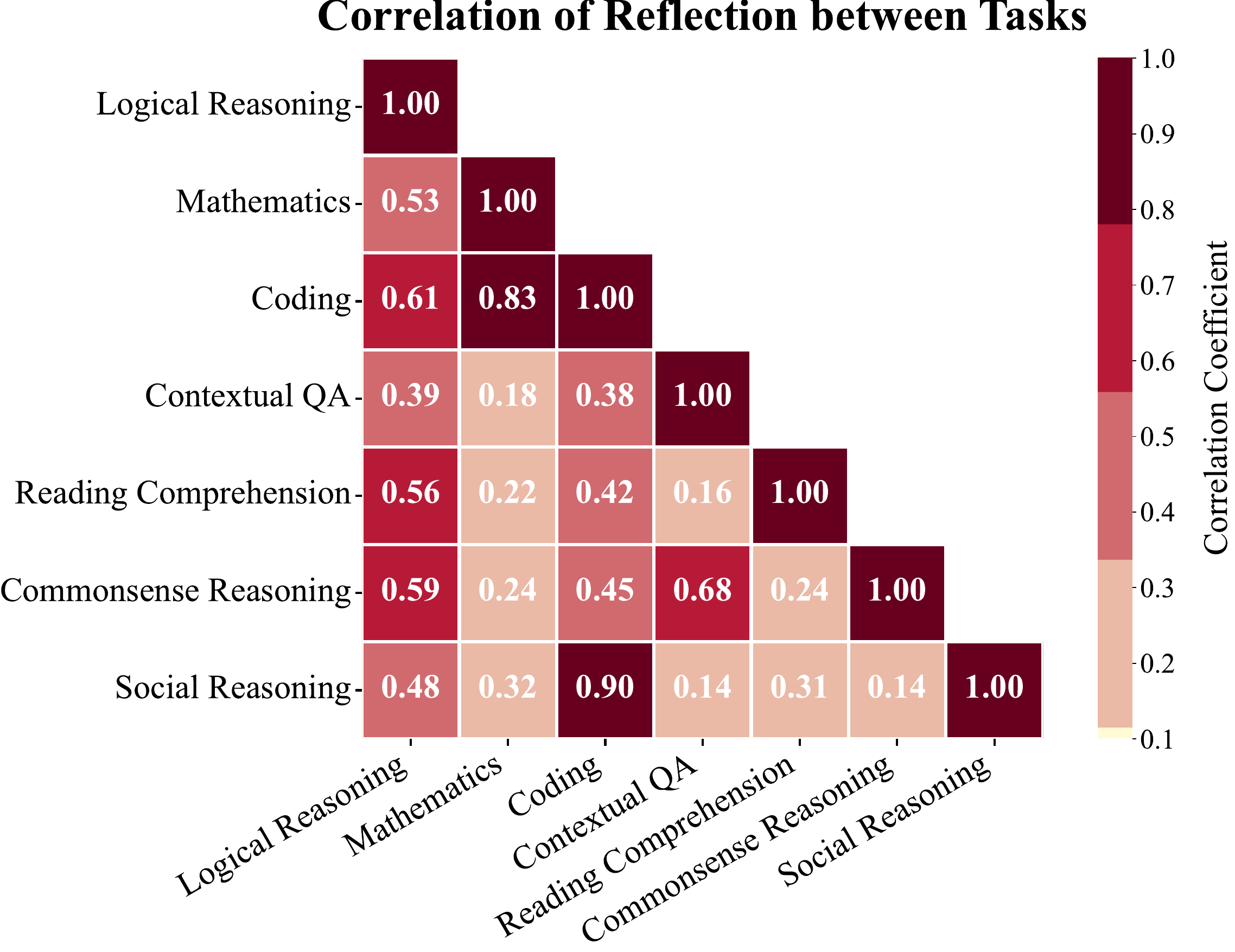}
    \caption{Correlation of reflection between each pair of tasks. We obtain the semantic representation for all reflections via the \texttt{Nv-Embed-v2} model \cite{lee2025nvembed} and calculate the Spearman correlation between each pair of tasks. The results are as follows: 1) Logical Reasoning has a \textcolor{reddd}{moderate correlation} with all tasks, which indicates that logic is a fundamental ability that supports other tasks; 2) Coding and Math have a \textcolor{redd}{high correlation}, implying that similar thinking patterns are required for handling math and coding problems; and 3) Commonsense Reasoning and Social Reasoning show \textcolor{redddd}{low correlation} (0.14), suggesting that these abilities might require different cognitive skills.}
    \label{fig:task_reflection_correlation}
\end{figure*}

\clearpage

\section{Prompts}
\label{appendix:prompts}

\subsection{Reflection generation for \ds}\label{appendix: rfl_prompt}
\begin{tcolorbox}[colback=gray!10, colframe=black, width=15.5cm,
                  arc=1mm, auto outer arc, boxrule=0.5pt,
                 ]
\textbf{Instruction:} Given the question and relevant context, you were unsuccessful in answering the question. As an advanced reasoning agent, you are required to enhance your incorrect solution and correctly answer the question based on self-reflection.\\
\\
Question: \textit{\{Question\}}\\
\\
Previous trial and your incorrect solution: \textit{\{Scratchpad\}}
\\
\\
Based on this information, please provide the following:

\# Stage 1: Verify the failed solution \\
\hspace*{0.5cm} 1-1. Analyze the failed solution by tracing and examining its execution with step-by-step verification.\\
\hspace*{0.5cm} 1-2. Quickly go through the failed solution without step-by-step verification.  \\

\# Stage 2: Locate errors and diagnose potential reasons \\
\hspace*{0.5cm} Identify specific steps where the errors occur and diagnose potential reasons. \\
\hspace*{0.5cm} 2-1. Review your calculation process to ensure that all the operations are accurate. \\
\hspace*{0.5cm} 2-2. Review your algorithm logic to ensure all steps follow the correct order. \\
\hspace*{0.5cm} 2-3. Review your solution to ensure it maintains logical coherence. \\
\hspace*{0.5cm} 2-4. Review your solution to check statements and conclusions for internal consistency. \\
\hspace*{0.5cm} 2-5. Review the context and requirements presented in the question to confirm that the response aligns with them. \\
\hspace*{0.5cm} 2-6. Review your solution to ensure that it is relevant to the question and addresses each aspect of the question. \\
\hspace*{0.5cm} 2-7. Review your solution to ensure it conforms to the required format and guidelines in a well-organized structure. \\
\hspace*{0.5cm} 2-8. Review your solution to ensure all provided facts are accurate and up to date. \\

\# Stage 3: Outline strategies and plans on error correction and mitigation \\
\hspace*{0.5cm} 3-1. Outline a high-level plan explaining how these changes will mitigate similar issues. \\
\hspace*{0.5cm} 3-2. Outline a low-level plan proposing specific strategies or corrections to address these issues.\\

Please follow the instructions without any additional introductory or concluding statements. Do not provide the answer directly. You will be punished to output more than 100 words.
\end{tcolorbox}

\subsection{Self-reflection for Reflector}
\begin{tcolorbox}[colback=gray!10, colframe=black, width=15.5cm,
                  arc=1mm, auto outer arc, boxrule=0.5pt,
                 ]
\textbf{Instruction:} You are an advanced reasoning agent that can improve based on self-reflection. You will be given a previous reasoning trial in which you were given a question to answer. You were unsuccessful in answering the question. In a few sentences, diagnose a possible reason for failure and devise a new, concise, high-level plan that aims to mitigate the same failure. Use complete sentences.\\
\\
Question: \textit{\{Question\}}\\
Previous trial and your incorrect solution: \textit{\{Scratchpad\}}
\end{tcolorbox}

\subsection{Reasoning for Generator}
\begin{tcolorbox}[colback=gray!10, colframe=black, width=15.5cm,
                  arc=1mm, auto outer arc, boxrule=0.5pt,
                 ]
\textbf{Instruction:} In this task, you are required to solve a question with interleaving Thought, Action, and Observation steps. Thought allows you to reason and analyze the current situation. Action calls the `Finish' function and fill in the answer in [ ] to finish the task after Thought. The observations will be provided to you automatically after you action.\\
\\
You can think step-by-step to reach the answer. Here are some examples:\\
\textit{\{Examples\}}\\
(END OF EXAMPLES)\\
\\
You are solving the following question: \textit{\{Question\}}\\
\\
Below is your progress so far:\\
(BEGIN)\\
\textit{\{Scratchpad\}}\\
(END)\\
\\
Please complete the current step.
\end{tcolorbox}

\subsection{Self-correct for Reflector}
\begin{tcolorbox}[colback=gray!10, colframe=black, width=15.5cm,
                  arc=1mm, auto outer arc, boxrule=0.5pt,
                 ]
\textbf{Instruction:} In this task, you are required to solve a question with interleaving Thought, Action, and Observation steps. Thought allows you to reason and analyze the current situation. Action calls the `Finish' function and fill in the answer in [] to finish the task after Thought. The observations will be provided to you automatically after you action.\\
\\
You can think step-by-step to reach the answer. Here are some examples:\\
\textit{\{Examples\}} \\
(END OF EXAMPLES)\\
\\
You are solving the following question: \textit{\{Question\}}\\
\\
Below is your previous reflection that helps to revise the incorrect solutions and correctly answer the question. It localizes the errors, summarizes the potential reasons for your failure and outlines a plan to mitigate the same failure:\\
\textit{\{Reflections\}}\\
\\
Below is your progress so far:\\
(BEGIN)\\
\textit{\{Scratchpad\}}\\
(END)\\
\\
Please complete the current step.
\end{tcolorbox}

\subsection{Self-reflection and correct in one stage}
\begin{tcolorbox}[colback=gray!10, colframe=black, width=15.5cm,
                  arc=1mm, auto outer arc, boxrule=0.5pt,
                 ]
\textbf{Instruction:} You are an advanced reasoning agent that can improve based on self-reflection. You will be given a previous reasoning trial in which you were given a question to answer. You were unsuccessful in answering the question. In a few sentences, diagnose a possible reason for failure and devise a new, concise, high-level plan that aims to mitigate the same failure. Use complete sentences.
\\
\\
Question: \textit{\{Question\}}
\\
\\
Previous trial and your incorrect solution: \textit{\{Scratchpad\}}
\\
\\
Based on your self-reflection, you can think step-by-step to generate a new answer to the question. Call the `Finish' function and fill in the answer in [ ] to finish the task.
\end{tcolorbox}

\subsection{Reasoning for direct QA for SFT}
\begin{tcolorbox}[colback=gray!10, colframe=black, width=15.5cm,
                  arc=1mm, auto outer arc, boxrule=0.5pt,
                 ]
\textbf{Instruction:} In this task, you are required to solve a question by generating the final answer directly.  Call the `Finish’ function and fill in the answer in [ ] to finish the task.
\\
Here are some examples: \\
\textit{\{Examples\}} \\
(END OF EXAMPLES)\\
\\
You are solving the following question: \textit{\{Question\}}
\end{tcolorbox}

\subsection{Reflection preference annotation by GPT}
\begin{tcolorbox}[colback=gray!10, colframe=black, width=15.5cm,
                  arc=1mm, auto outer arc, boxrule=0.5pt,
                 ]
\textbf{Instruction:} You are a helpful assistant in evaluating the quality of reflections on an unsuccessful attempt to answer a question.
\\

You will be provided with:
\\

Question:  \textit{\{Question\}}

Groundtruth to the question:  \textit{\{Answer\}}

Initial student's chain of thought and answer:  \textit{\{Scratchpad\}}

Student A's reflection:  \textit{\{Reflections 1\}}

Student B's reflection:  \textit{\{Reflections 2\}} \\

Student A and Student B have both reflected on the initial student's unsuccessful attempt to answer the question. Above are their reflections that diagnose possible reasons for failure or devise a better plan to mitigate the same failure. Please determine which student's reflection is better. \\

Your response should be either "Student A" or "Student B" without providing any explanation or other words for your choice. \\

Do NOT say both/neither are good.
\end{tcolorbox}

\subsection{Heuristic Error Constituent Annotation for Reflection}
\begin{tcolorbox}[colback=gray!10, colframe=black, width=15.5cm,
                  arc=1mm, auto outer arc, boxrule=0.5pt,
                 ]
\textbf{Instruction:} You are a professional data annotator specializing in reasoning and chain-of-thought rationale analysis. Your task is to analyze the thought process provided and identify any fine-grained labels based on the given reflection and error taxonomy.
\\\\
\# Definitions
\\
- Thought: The reasoning steps taken by a human or model to arrive at an answer.
\\
- Reflection: The self-reflection of the human or model on the reasoning thought process.
\\\\
\# Error Taxonomy
\\
1. Mathematical Errors
\\
$\bullet$  1-1. Calculation Error
   \\
$\bullet$  1-2. Algorithm Error
\\\\
2. Logic and Reasoning Errors
\\
$\bullet$  2-1. Flawed Rationale Error
   \\
$\bullet$  2-2. Internal Inconsistency
\\\\
3. Instruction Violation
\\
$\bullet$  3-1. Context Misinterpretation
   \\
$\bullet$  3-2. Incomplete or Irrelevant Response
   \\
$\bullet$  3-3. Format Discrepancy
\\\\
4. Factual Errors
\\
$\bullet$  4-1. Factual Errors
\\\\
5. No Errors
\\
$\bullet$  5-1. No Errors Detected
\\\\
\# Input
\\
- Question: \textit{\{question\}}
\\
- Thought: \textit{\{thought\}}
\\
- Reflection: \textit{\{reflection\}}
\\\\
\# Output
\\
- Labels: [Error Type(s) Assigned]
\\
- Rationale: [Explanation for label assignment, with specific examples]
\end{tcolorbox}

\end{document}